# Efficiency Improvement of Measurement Pose Selection Techniques in Robot Calibration

Y.Wu[1,2], A. Klimchik[1,2], A. Pashkevich[1,2], S.Caro[2], B.Furet[3]

[1]*Ecole des Mines de Nantes,44307, Nantes, France*
[2]*Institut de Recherche en Communications et Cybernétique de Nantes (IRCCyN), 44321, Nantes, France*
[3]*Université de Nantes, 44322, Nantes, France*
*(e-mail:{ yier.wu, alexandr,klimchik, anatol,pashkevich} @mines-nantes.fr,*
*stephane.caro@irccyn.ec-nantes.fr,  benoit.furet@univ-nantes.fr)*

**Abstract:** The paper deals with the design of experiments for manipulator geometric and elastostatic calibration based on the test-pose approach. The main attention is paid to the efficiency improvement of numerical techniques employed in the selection of optimal measurement poses for calibration experiments. The advantages of the developed technique are illustrated by simulation examples that deal with the geometric calibration of the industrial robot of serial architecture.

*Keywords*: robot calibration, measurement pose selection, test-pose approach

## 1. INTRODUCTION

In the usual engineering practice, the accuracy of a manipulator depends on a number of factors. Usually in robotics, the geometric and elastostatic errors are the most significant ones. Their influence on the robot positioning accuracy highly depends on the manipulator configuration and essentially differs throughout the workspace. To achieve good accuracy in all working points, adequate geometric and stiffness models are required. While the model structure is usually well known, the identification of the model parameters (calibration) is rather time consuming and requires essential experimental work. For this reason, optimal selection of measurement poses for robot calibration is an important problem, which is still in the focus of numerous research papers (Daney 2002, Sun 2008).

At present, the main activity in this area is concentrated around the geometric calibration (Khalil 2002). On the other hand, the elastostatic calibration which is also very important for many applications (such as precise machining) has attracted less attention of the researchers (Meggiolaro 2005). However, for both of these calibration procedures, the problem of measurement pose selection is one of the key issues allowing to reduce essentially the measurement error impact (Klimchik 2011). At first sight, this problem can be solved using well known results from the classical design of experiments theory. However, because of the specificity and nonlinearity of the manipulator geometric and elastostatic models, the problem solution is not so obvious. The main difficulties here are in the area of definition of a reasonable optimality criterion (which has clear engineering sense) and also in efficient solution of the relevant optimization problem, which has rather high dimension.

Among related works, it is worth mentioning several papers. The majority of the measurement pose selection techniques relies on the optimization of some functions depending on the singular values of the identification Jacobian. For example, Zhuang used genetic algorithm for minimization of the condition number of this matrix (Zhuang 1996). In other work (Daney 2005), to decrease the sensitivity to local minima, Daney developed the local convergence method and Tabu search technique based on the observability index. However, the performance measures used in these works are rather abstract and are not directly related to the robot accuracy. Besides, the related objective functions are very difficult for the optimization due to existence of a number of local minima. To find the global one, heuristic search is usually used as the numerical algorithms, which often require tedious computations. All these motivate the research direction of this work.

In this paper, the problem of optimal design of calibration experiments is studied for the case of robot manipulator of serial architecture. In contrast to other works, the optimization problem related to measurement pose selection is formulated using the proposed performance measure (test-pose approach), which has clear physical meaning and is directly related to robot accuracy. The main attention is paid to the efficiency improvement of the related numerical routines.

## 2. PROBLEM STATEMENT

*2.1 Geometric and Elastostatic Models of Manipulator*

In industrial robot controllers, the end-effector position of the manipulator is usually computed using the geometric model. For some specific applications, such as high-speed machining that generate essential external loading, the elastostatic model should be also used. However, in practice, the robot geometric parameters essentially differ from the nominal values declared in technical specifications and vary from one robot to another. In addition, elastostatic parameters of the manipulator are not provided by the robot manufacturers and can be identified from the experiments only. So, the manipulator model parameter identification is an important step in practical application of industrial robots.

The manipulator geometric model provides the position/orientation of robot end-effector as a function of the joint variables and its inherent parameters. This model is usually presented as a product of homogeneous transformation matrices, which after some transformations can be presented as the vector function

$$\mathbf{p} = g(\mathbf{q}, \mathbf{\Pi}) \quad (1)$$

where vector $\mathbf{p}$ denotes the end-effector position, vector $\mathbf{q}$ aggregates all joint angles and $\mathbf{\Pi}$ are the vector of unknown parameters to be identified. These unknowns differ with the applied parameterization methods in robot geometric modelling, such as the classical Denavit and Hartenberg approach and its modified version (Khalil 1986). In this paper, there are considered the most essential components of the vector $\mathbf{\Pi}$, which are the deviations of the robot link lengths $\Delta l_i$ and the offsets $\Delta q_i$ in the actuated joints. Since the deviations of geometrical parameters $\Delta \mathbf{\Pi}$ are usually relatively small, calibration usually relies on the linearized model

$$\mathbf{p} = g(\mathbf{q}, \mathbf{\Pi}_0) + \mathbf{J}_g(\mathbf{q}, \mathbf{\Pi}_0) \Delta \mathbf{\Pi} \quad (2)$$

which includes the conventional geometric Jacobian $\mathbf{J}_g(\mathbf{q}, \mathbf{\Pi}_0) = \partial g(\mathbf{q}, \mathbf{\Pi}_0)/\partial \mathbf{\Pi}$ computed for the nominal geometric parameters $\mathbf{\Pi}_0$.

The elastostatic properties of a serial robotic manipulator represent its resistance to deformations caused by external forces/torques and are usually described by the Cartesian stiffness matrix $\mathbf{K}_C$, which is computed as

$$\mathbf{K}_C = \mathbf{J}_\theta^{-T} \mathbf{K}_\theta \mathbf{J}_\theta^{-1} \quad (3)$$

where $\mathbf{K}_\theta$ is a diagonal matrix that aggregates the joint stiffness values (that are the unknowns to be identified) and $\mathbf{J}_\theta$ is the corresponding elastostatic Jacobian. This model can be derived using the virtual joint method, which describes all elastostatic properties of compliant elements by localized virtual springs located in the actuated joints (Salisbury 1980). Using the Cartesian stiffness matrix, the elastostatic model (or force-deflection relation) can be expressed as

$$\mathbf{w} = \mathbf{J}_\theta^{-T} \mathbf{K}_\theta \mathbf{J}_\theta^{-1} \cdot \Delta \mathbf{p} \quad (4)$$

where $\Delta \mathbf{p}$ is the position deflection at the robot end-effector caused by the external wrench $\mathbf{w}$, which integrates both the external force and torque. This linear relation can be further used for the calibration where the desired parameters to be identified are the components of matrix $\mathbf{K}_\theta$.

In the frame of this work, several assumptions concerning calibration of these models are accepted:

*A1:* For the geometric calibration, *each calibration experiment produces two vectors* $\{\Delta \mathbf{p}_i, \mathbf{q}_i\}$, which define the robot end-effector displacements and corresponding joint angles.

The linear relation between the errors in geometric parameters and the end-effector position deviations can be written as

$$\Delta \mathbf{p}_i = \mathbf{J}_g(\mathbf{q}_i) \Delta \mathbf{\Pi} \quad (5)$$

where $\mathbf{J}_g(\mathbf{q}_i)$ is the Jacobian matrix that depends on manipulator configuration $\mathbf{q}_i$ and vector $\Delta \mathbf{\Pi}$ collects the unknown parameters to be identified.

*A2:* For the elastostatic calibration, *each calibration experiment produces three vectors* $\{\Delta \mathbf{p}_i, \mathbf{q}_i, \mathbf{w}_i\}$, where $\mathbf{w}_i$ defines the applied forces and torques.

In accordance with (Pashkevich 2011), the corresponding mapping from the external wrench space to the end-effector deflection space can be expressed as

$$\Delta \mathbf{p}_i = \mathbf{J}_\theta(\mathbf{q}_i) \mathbf{k}_\theta \mathbf{J}_\theta^T(\mathbf{q}_i) \mathbf{w}_i \quad (6)$$

where $\mathbf{k}_\theta$ is a matrix that aggregates the unknown compliance parameters $\{k_1, ..., k_n\}$ to be identified.

Hence, the calibration experiments provide the set of vectors $\{\Delta \mathbf{p}_i, \mathbf{q}_i\}$ and $\{\Delta \mathbf{p}_i, \mathbf{q}_i, \mathbf{w}_i\}$ that allow us to estimate the deviations in geometric parameters $\Delta \mathbf{\Pi}$ (compared to the nominal values) and absolute values of the elastostatic parameters included in the diagonal matrix $\mathbf{k}_\theta$.

### 2.2 Identification of the Model Parameters

The problem of parameter identification of the robot manipulator can be treated as the best fitting of the experimental data by corresponding models. These data are measured under several assumptions concerning the measurement equipment:

*A3:* The calibration relies on the *measurements of the end-effector position only* (Cartesian coordinates $\{p_x, p_y, p_z\}$).

*A4: The measurements errors* $\boldsymbol{\varepsilon}_i$ accommodated in each measurement of end-effector position are treated as independent identically distributed random values with zero expectation and standard deviation $\sigma$.

For computational convenience and taking into account the influence of measurement errors, the geometric and elastostatic models described by separate linear equations (5) and (6) can be expressed in the following integrated form

$$\Delta \mathbf{p}_i = \mathbf{B}(\mathbf{q}_i) \cdot \Delta \mathbf{X} + \boldsymbol{\varepsilon}_i \quad (7)$$

where $\Delta \mathbf{X} = \{\Delta \mathbf{\Pi}, \mathbf{k}_\theta\}$ collects all unknown parameters (both geometric and elastostatic ones), and the matrix $\mathbf{B}$ varies depending on different calibration cases

$$\mathbf{B}_{3m \times 3n} = \begin{cases} [\mathbf{J}_{3m \times 2n} \quad \mathbf{0}_{3m \times n}], & \text{for geometric parameters} \\ [\mathbf{0}_{3m \times 2n} \quad \mathbf{A}_{3m \times n}], & \text{for elastostatic parameters} \\ [\mathbf{J}_{3m \times 2n} \quad \mathbf{A}_{3m \times n}], & \text{for both parameters} \end{cases} \quad (8)$$

where $\mathbf{J}$ is the Jacobian matrix that can be obtained by differentiating the manipulator geometric model with respect to the desired parameters; and matrix $\mathbf{A}$ can be computed as

$$\mathbf{A}(\mathbf{q}_i, \mathbf{w}_i) = [\mathbf{J}_1(\mathbf{q}_i) \mathbf{J}_1^T(\mathbf{q}_i) \mathbf{w}_i, ..., \mathbf{J}_n(\mathbf{q}_i) \mathbf{J}_n^T(\mathbf{q}_i) \mathbf{w}_i] \quad (9)$$

where $\mathbf{J}_n(\mathbf{q}_i)$ is the $n$-th column vector of the Jacobian matrix for the $i$-th experiment. $n$ is the number of joints.

Using usual approach adopted in the identification theory, the estimated unknown parameters $\Delta \hat{\mathbf{X}} = \{\Delta \hat{\mathbf{\Pi}}, \hat{\mathbf{k}}_\theta\}$ can be obtained using the least square method, which yields

$$\Delta \hat{\mathbf{X}} = \left( \sum_{i=1}^{m} \mathbf{B}_i^T \mathbf{B}_i \right)^{-1} \cdot \left( \sum_{i=1}^{m} \mathbf{B}_i^T \Delta \mathbf{p}_i \right) \quad (10)$$

Using this expression, it can be proved that the covariance matrix for the identification errors in the parameters $\Delta \mathbf{X}$ can be computed as

$$\text{cov}(\Delta \hat{\mathbf{X}}) = \sigma^2 \left( \sum_{i=1}^{m} \mathbf{B}_i^T \mathbf{B}_i \right)^{-1} \quad (11)$$

where $\sigma$ is the standard deviation (s.t.d.) of the measurement errors. Hence, the impact of the measurement errors on the parameter identification accuracy is defined by the matrix sum $\sum_{i=1}^{m} \mathbf{B}_i^T \mathbf{B}_i$ that is also called the information matrix.

It is obvious that, from practical point of view, the covariance matrix should be as small as possible. However, strict mathematical definition of this notion is not trivial and a number of different approaches are proposed in literature. In most of the related works, the optimal measurement poses are obtained based on minimization of the covariance matrix norm (Atkinson 1992). This approach may provide a solution, which does not guarantee the best position accuracy for typical manipulator configurations defined by the manufacturing process. Thus, here it is proposed the industry-oriented performance measure, $\rho_0^2$, which is defined as the mean square error in the end-effector position after compensation.

To develop this approach, let us introduce several definitions:

**D1:** *Plan of experiments* is a set of robot configurations $\mathbf{Q}$ and corresponding external loadings $\mathbf{W}$ that are used for the measurements of the end-effector displacements and further identification of the desired parameters.

**D2.** The *accuracy of the error compensation* $\rho_0$ is the distance between the desired end-effector position and its real position achieved after application of error compensation technique.

**D3.** The *manipulator test-pose* is one or set of robot configurations $\mathbf{Q}_0$ and corresponding external loadings $\mathbf{W}_0$ for which it is required to achieve the best error compensation (i.e. $\rho_0^2 \to \min$).

In the frame of the adopted notations, the distance defining the error compensation accuracy can be computed as

$$\delta \mathbf{p} = \mathbf{B}_0 (\Delta \hat{\mathbf{X}} - \Delta \mathbf{X}) \quad (12)$$

where the vectors $\Delta \mathbf{X}$ and $\Delta \hat{\mathbf{X}}$ are the true parameters values and their estimates, respectively. Matrix $\mathbf{B}_0$ corresponds to the test pose (see expressions (8)). Further, taking into account that $\delta \mathbf{p}$ is a function of the unbiased random variables $\{\boldsymbol{\varepsilon}_1, ..., \boldsymbol{\varepsilon}_m\}$, it can be easily proved that the expectation $\text{E}(\delta \mathbf{p}) = 0$. Besides, the variance can be expressed as

$$\text{Var}(\delta \mathbf{p}) = \text{E} \left( \delta \mathbf{X}^T \mathbf{B}_0^T \mathbf{B}_0 \delta \mathbf{X} \right) \quad (13)$$

where $\delta \mathbf{X} = \Delta \mathbf{X} - \Delta \hat{\mathbf{X}}$ is the difference between the estimated and true values of the parameters. Expression. (13) can be rewritten as $\text{trace}(\mathbf{B}_0 \text{E}(\delta \mathbf{X} \delta \mathbf{X}^T) \mathbf{B}_0^T)$ and after relevant transformations in accordance with (10), (11), yields the desired expression for the compensation accuracy

$$\rho_0^2 = \sigma^2 \cdot \text{trace} \left( \mathbf{B}_0 \left( \sum_{i=1}^{m} \mathbf{B}(\mathbf{q}_i)^T \mathbf{B}(\mathbf{q}_i) \right)^{-1} \mathbf{B}_0^T \right) \quad (14)$$

As follows from this expression, the proposed performance measure can be treated as the weighted trace of the covariance matrix (11), where the weighting coefficients are computed using the test pose.

Hence, the identification quality (evaluated via the error compensation accuracy) is completely defined by the set of matrices $\{\mathbf{B}_1, ..., \mathbf{B}_m\}$ that depend on the manipulator configurations $\{\mathbf{q}_1, ..., \mathbf{q}_m\}$. Optimal selection of these configurations will be in the focus of next Subsection.

*2.3 Problem of the Measurement Poses Selection*

Based on the performance measure presented in the previous Subsection, the corresponding optimization problem of the measurement pose selection can be defined as

$$\text{trace} \left( \mathbf{B}_0 \left( \sum_{i=1}^{m} \mathbf{B}(\mathbf{q}_i)^T \mathbf{B}(\mathbf{q}_i) \right)^{-1} \mathbf{B}_0^T \right) \to \min_{\mathbf{q}_i, \mathbf{w}_i} \quad (15)$$

$$\text{subject to} \quad \mathbf{C}_i(\mathbf{q}_i, \mathbf{w}_i) \leq 0, \quad i = \overline{1, r}$$

Here, the matrices $\mathbf{C}_i(\mathbf{q}_i, \mathbf{w}_i)$ describe some constraints, which should be taken into account while solving optimization problem. These constraints are imposed by the work-cell design particularities and usually include the manipulator joint limits, the work-cell space limits, measurement equipment limitations, etc. It should be also mentioned that some constraints are imposed to avoid collisions between the work-cell components and the manipulator. Besides, some directions of the applied loading are preferable for the reason of practical implementation.

It should be mentioned that the component of the matrices $\mathbf{C}_i$ vary with different calibration cases and may include some very specific constraints. For instance, for the case of elastostatic calibration, they can be expressed as

$$\mathbf{C}_1 = \begin{bmatrix} q_i - q_i^{\max} \\ q_i^{\min} - q_i \end{bmatrix}, \quad \mathbf{C}_3 = \begin{bmatrix} p_z^{\min} - p_z \\ r^{\min} - r \\ \varphi^{\min} - |\varphi| \end{bmatrix}, \quad \mathbf{C}_4 = \begin{bmatrix} \mathbf{p}_i - \mathbf{p}_i^{\max} \\ \mathbf{p}_i^{\min} - \mathbf{p}_i \end{bmatrix} \quad (16)$$

$$\mathbf{C}_2 = \|\mathbf{F}^i\| - F_{\max}$$

where $q_i^{\min}$ and $q_i^{\max}$ are the joint limits, $F_{\max}$ is the robot maximum payload, $p_z^{\min}$ is the minimum height between the end-point of the calibration tool and the work-cell floor, $r^{\min}$ is the minimum radius to avoid collisions between the applied loading and robot body, $\varphi^{\min}$ is the minimum angle between the direction of calibration tool and z-axis of robot base frame to ensure that the vertical loading can be applied, $\mathbf{p}_i^{\min}$ and $\mathbf{p}_i^{\max}$ are the boundaries of work-cell space. For the case of geometric calibration, the problem of applying external loading does not exist. So, $\mathbf{C}_2$, $\mathbf{C}_3$ are zero matrices, while $\mathbf{C}_1$, $\mathbf{C}_4$ remain the same as in elastostatic calibration.

The procedure of solving such an optimization problem could be very tedious for the case when numerous measurement configurations are required for the calibration experiment. For this reason, the problem of interest is to find reasonable number of different measurement configurations and to improve the efficiency of optimization routines employed in the measurement pose selection.

## 3. MEASUREMENT POSE SELECTION TECHNIQUES

To solve the above define problem, several techniques can be applied. This section presents the analysis and propose some approaches allowing to obtain acceptable solution in reasonable time. The main difficulties here are related with a large number of variables and complex behaviour of the objective function that has many local minima.

### 3.1 Using Conventional Optimization Techniques

The simplest way to solve this problem is to apply conventional optimization techniques incorporated in commercial mathematical software. It is clear that straightforward search with regular grid is non-acceptable here because of high complexity and enormous number of solutions to be compared. For this reason, three other algorithms have been examined: (i) random search, (ii) gradient search, and (iii) genetic algorithm. Their comparison study is presented below and summarized in Tables 1 and 2, where two criteria have been used: computational time and the ability to find optimal solution (evaluated via $\rho_0$, the manipulator accuracy after calibration). For all computational experiments, it was assumed that the s.t.d. of the measurement errors is 0.03mm.

The benchmark example deals with the calibration experiments design for 6-dof industrial manipulator KUKA KR270, whose nominal parameters can be found on the manufacturer website (www.kuka.com). The robot has a serial architecture with six actuated revolute joints, so 24 independent geometric parameters should be identified in general case. But for this example, to reduce computational efforts and evaluate the algorithm capability before applying to the problem of real dimension, only nine of the most essential parameters were identified (which have major impact on the positioning accuracy). This allowed us to obtain realistic assessments of the conventional optimization techniques capabilities with respect to the considered problem where the number of design variables is high enough (72 for 12 configurations).

The first of the examined algorithm (i) is based on the straightforward selection of the best solution from the set of ones generated in a random way. For this study, 10,000 solutions were generated for different numbers of measurement configurations $m \in \{3, 4, 6, 12\}$. As follows from the obtained results (see Tables 1 and 2), this algorithm is very fast and requires less than 2 minutes to find the best solution. However, this solution is essentially worse than the optimal one (by 15-30%).

The second algorithm (ii) employs the gradient search with built-in numerical evaluation of the derivatives that is available in Matlab. The starting points were generated randomly and, to avoid convergence to the local minima, the optimization search has been repeated 5000 times (starting from different points). In this case, it has been obtained the best result in terms of the desired objective $\rho_0$, but computational cost was very high (it can overcome a hundred of hours). So, this technique is hardly acceptable in practice. It is worth mentioning that reduction of the iteration number is rather dangerous here, because there are a number of local minima that the algorithm can converge to (see Table 1 that includes the minimum, maximum and average values of $\rho_0$ obtained for random starting points). Moreover, as follows from our experience, 5000 iterations are also not enough here.

Table 1. Efficiency of conventional optimization techniques

| Algorithm | | Number of poses | | | |
|---|---|---|---|---|---|
| | | $m=3$ | $m=4$ | $m=6$ | $m=12$ |
| Random Search | $\rho_0^{\min}$ [mm] | 0.0825 | 0.0607 | 0.0519 | 0.0360 |
| | $\rho_0^{\text{mean}}$ [mm] | 1.9155 | 0.1939 | 0.0905 | 0.0500 |
| | $\rho_0^{\max}$ [mm] | 29.878 | 14.9746 | 0.5608 | 0.0804 |
| Gradient Search | $\rho_0^{\min}$ [mm] | 0.0637 | 0.0521 | 0.0426 | 0.0302 |
| | $\rho_0^{\text{mean}}$ [mm] | 0.0862 | 0.0610 | 0.0477 | 0.0335 |
| | $\rho_0^{\max}$ [mm] | 0.6534 | 0.2333 | 0.1319 | 0.0681 |
| Genetic Algorithm | $\rho_0^{\min}$ [mm] | 0.0638 | 0.0521 | 0.0427 | 0.0301 |
| | $\rho_0^{\text{mean}}$ [mm] | 0.0689 | 0.0529 | 0.0433 | 0.0305 |
| | $\rho_0^{\max}$ [mm] | 0.0802 | 0.0615 | 0.0446 | 0.0309 |

The third of the examined techniques (iii) applies genetic algorithm (GA) that is based on adaptive heuristic search. The optimization has been carried out for 100 times with population size 50 and 20 generations (initial populations were randomly generated). For illustrative purposes, Fig.1 presents the efficiency of this algorithm for selection of three optimal measurement configurations. It shows the algorithm convergence as well as divergence of the optimal solutions with respect to computing time. As follows from this figure, the optimization results are highly sensitive to the selection of initial population. In particular, the diversity of the optimal solutions got from sequential GA runs is about 25%. So, to achieve the global minimum, the GA should be repeated many times, which leads to essential increase of the computational efforts (more than 6 hours of computations for the considered example). However, compared to gradient search, GA provides acceptable accuracy (only 2% worse) while the computational time is 4 times less.

Table 2. Computational time of examined algorithms

| Algorithm | Number of poses | | | |
|---|---|---|---|---|
| | $m=3$ | $m=4$ | $m=6$ | $m=12$ |
| Random Search | 41s | 47s | 1min | 1.7min |
| Gradient Search | 24.2h | 37.5h | 56.3h | 103.6h |
| Genetic Algorithm | 6.5h | 8.3h | 10.5h | 15.4h |

As follows from the obtained results, the random search is rather fast but inefficient here, since it may produce non-acceptable solutions. In contrast, the gradient search is able to find the global minimum provided that it is repeated many times with different starting points. As a compromise, the GA provides intermediate results in terms of accuracy and computational time. However, for problems of the real industrial size, the performances of the GA are also not sufficient. For this reason, the following Subsections are devoted to the improvement of the numerical optimization techniques employed in selection of optimal measurement poses.

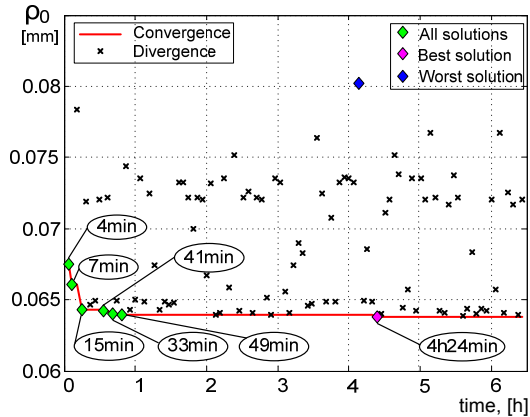

Figure 1. Efficiency of GA for selection of three measurement poses (population size 50, 20 generations)

*3.2 Applying Parallel Computing*

Since the considered problem requires numerous repetitions of the optimization with different initial values, applying parallel computing looks attractive to speed up the design process and to take advantage of multi-core architecture available in modern computers.

To evaluate benefits of the parallel computing, the same benchmark example has been considered and two algorithms have been examined: (ii)' parallel gradient search, and (iii)' parallel GA with the same parameter settings. The computations were carried out on the workstation with 12 cores. The obtained results are presented in Table 3, which gives the computational time for different number of measurement poses (the attained value of the objective function $\rho_0$ is very close to those presented in Table 1).

The obtained results are quite expected and confirm essential reduction of computational efforts. For both optimization methods, the consumed time has been decreased by the factor of 10-12 (compared to the results in Table 2). However, it is not enough yet to solve the problem of real industrial size, where several dozen of parameters should be identified (instead of nine in the benchmark example).

Table 3. Computational time of examined algorithms using parallel computing

| Algorithm | Number of poses | | | |
| --- | --- | --- | --- | --- |
| | $m=3$ | $m=4$ | $m=6$ | $m=12$ |
| Parallel Gradient Search | 2.1h | 3.2h | 4.9h | 8.9h |
| Parallel Genetic Algorithm | 36min | 41min | 52min | 1.5h |

*3.3 Using Hybrid Approach*

To take the advantages of both examined algorithms and efficiency of the parallel implementation, a hybrid technique has been developed. It should be mentioned that some software packages (Matlab, etc.) already implement this idea and use the final solution from GA as the initial point of gradient search. However, since the randomly generated initial populations in GA may cause high diversity of the optimal solutions, the selection of these initial values is also an important issue. For this reason, the embedded hybrid option in GA cannot be directly used and requires additional modifications.

To improve the efficiency of the existing technique, the starting point selection strategy for the gradient search has been modified. To ensure better convergence to the global minimum, it has been proposed to use the best half of final points obtained from GA as the starting points for the gradient search. From our point of view, it ensures better diversity of the starting points and allows to avoid convergence to the local minima.

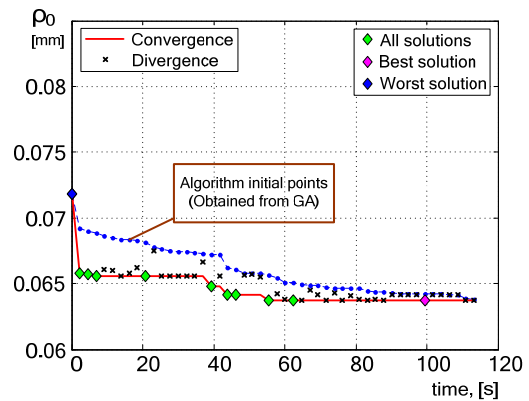

Figure 2. Efficiency of the hybrid approach for selection of three measurement poses

The proposed modification has been evaluated using the same benchmark example. For comparison purposes, Fig 2. presents the convergence of the hybrid method for the problem of optimal selection of three measurement configurations studied in the previous Subsection (see Fig. 1). It shows the initial points (obtained from GA), optimal solutions as well as the solution improvement with respect to time. As follows from the figure, the hybrid algorithm can converge much faster, but if the number of measurement poses is increased up to 12, the computational time is over 1.6 hour that is still unacceptable for industry.

*3.4 New Approach: Reduction of Problem Dimension*

Generally, as follows from the identification theory, the only way to improve calibration accuracy is to increase the number of measurements (provided that the reduction of the measurement errors is not possible). However, for the manipulator calibration problem, each measurement is associated with a certain robot configuration that also has influence on the final accuracy. It is clear that the best result is achieved if all measurement poses are different and have been optimized during the calibration experiment planning.

On the other hand, as follows from our experiences, the diversity of the measurement poses does not contribute significantly to the accuracy improvement if $m$ is high enough. This allows us to propose an alternative which uses the same measurement configurations several times (allowing to simplify and speed up the measurements). This approach will be further referred as "reduction of problem dimension".

To explain the proposed approach in more details, let us assume that the problem of the optimal pose selection has been solved for the number of configurations that is equal to $m$, and the obtained calibration plan ensures the positioning accuracy $\rho_0^m$. Using these notations, let us evaluate the calibration accuracy for two alternative strategies that employ larger number of experiments $km$:

**Strategy #1** (*conventional*): the measurement poses are found from the full-scale optimization of size $km$.

**Strategy #2** (*proposed*): the measurement poses are obtained by simple repetition the configurations got from the low-dimensional optimization problem of size $m$.

It is clear that the calibration accuracy $\rho_0^{km}$ for strategy #1 is better than the accuracy corresponding to the strategy #2 that can be expressed as $\rho_0^m/\sqrt{k}$. However, as follows from our study, this difference is not high if $m$ is larger than 3. This allows us to essentially reduce the size of optimization problem employed in the optimal selection of measurement poses without significant impact on the positioning accuracy.

To demonstrate the validity of the proposed approach, the benchmark example has been solved using strategies #1 and #2 assuming that the total number of measurements is equal to 12 (i.e. using different factorizations such as $12\times1$, $6\times2$, $4\times3$, $3\times4$). Relevant results are presented in Table 4 (see the last line). As follows from them, the factorization $12\times1$ where all measurement poses are different is only 6% better compared to the factorization $3\times4$ where measurements are repeated 4 times in 3 different configurations. At the same time the factorizations $6\times2$ and $4\times3$ give almost the same results as the optimal factorizations $12\times1$. On the other hand, the computational time of the optimal pose generation for $m=3$ is much lower than for $m=12$. This demonstrates the efficiency of the proposed approach and justify its validity.

Table 4. Calibration accuracy $\rho_0$ for different factorizations of the experiment number $m = km_0$

| Number of measurements | | Number of different poses | | | |
|---|---|---|---|---|---|
| | | $m_0 = 3$ | $m_0 = 4$ | $m_0 = 6$ | $m_0 = 12$ |
| $m = 3$ | $\rho_0^{min}$ | 0.0637 (3×1) | | | |
| $m = 4$ | $\rho_0^{min}$ | | 0.0521 (4×1) | | |
| $m = 6$ | $\rho_0^{min}$ | 0.0450 (3×2) | | 0.0426 (6×1) | |
| $m = 12$ | $\rho_0^{min}$ | 0.0319 (3×4) | 0.0301 (4×3) | 0.0301 (6×2) | 0.0301 (12×1) |
| Computing time | | 38min | 45min | 56min | 1.6h |

Hence, it can be concluded that repeating experiments with optimal plans obtained for the lower number of configurations provides almost the same performance as "full-dimensional" optimal plan. Obviously, this reduction of the measurement pose number is very attractive for the engineering practice.

## 4. CONCLUSIONS

The paper presents a new technique for optimal selection of measurement poses in robot calibration. In contrast to other works, it is proposed to evaluate the quality of calibration plans via the manipulator positioning accuracy for a given test pose, and to reduce number of design variables in the related optimization problem by means of repeating measurements with lower number of configurations. This technique allows us to essentially reduce the computational time for solving the problem of real industrial size. The advantages of the developed technique were confirmed by a simulation example, where the proposed approach permitted to decrease the computing time by more than 10 times while losing only 6% of manipulator positioning accuracy. Future work in this direction will deal with the efficiency improvement of the manipulator elastostatic calibration.

## ACKNOWLEGMENT

The authors would like to acknowledge the financial support of the French "Agence Nationale de la Recherche"(Project ANR-2010-SEGI-003-02-COROUSSO), France.